%% file: main.tex
\documentclass[sigconf]{acmart}
\usepackage[symbol]{footmisc}

\AtBeginDocument{%
 }

\copyrightyear{2025}
\acmYear{2025}
\setcopyright{cc}
\setcctype{by}
\acmConference[KDD '25]{Proceedings of the 31st ACM SIGKDD Conference on
Knowledge Discovery and Data Mining V.2}{August 3--7, 2025}{Toronto, ON,
Canada}
\acmBooktitle{Proceedings of the 31st ACM SIGKDD Conference on Knowledge
Discovery and Data Mining V.2 (KDD '25), August 3--7, 2025, Toronto, ON,
Canada}
\acmDOI{10.1145/3711896.3736557}
\acmISBN{979-8-4007-1454-2/2025/08}

\begin{document}

\title{A Survey on Retrieval And Structuring Augmented Generation with Large Language Models}




\settopmatter{authorsperrow=4}

\author{Pengcheng Jiang*}
\affiliation{%
  \institution{University of Illinois Urbana-Champaign}
  \city{Urbana, IL}
  \country{USA}
  \institution{pj20@illinois.edu}
}

\author{Siru Ouyang*}
\affiliation{%
  \institution{University of Illinois Urbana-Champaign}
  \city{Urbana, IL}
  \country{USA}
  \institution{siruo2@illinois.edu}
}

\author{Yizhu Jiao}
\affiliation{%
  \institution{University of Illinois Urbana-Champaign}
  \city{Urbana, IL}
  \country{USA}
  \institution{yizhuj2@illinois.edu}
}

\author{Ming Zhong}
\affiliation{%
  \institution{University of Illinois Urbana-Champaign}
  \city{Urbana, IL}
  \country{USA}
  \institution{mingz5@illinois.edu}
}

\author{Runchu Tian}
\affiliation{%
  \institution{University of Illinois Urbana-Champaign}
  \city{Urbana, IL}
  \country{USA}
  \institution{runchut2@illinois.edu}
}

\author{Jiawei Han}
\affiliation{%
  \institution{University of Illinois Urbana-Champaign}
  \city{Urbana, IL}
  \country{USA}
  \institution{hanj@illinois.edu}
}

\renewcommand{\shortauthors}{Pengcheng Jiang et al.}


\input{sections/0-abstract}



\maketitle

\input{sections/1-introduction}
\input{sections/2-background}

\input{sections/3-retrieval}
\input{sections/4-structuring}
\input{sections/5-ras-app}

\input{sections/6-future}
\input{sections/7-conclusion}

\section*{Acknowledgment}
Research was supported in part by National Science Foundation IIS-19-56151, the Molecule Maker Lab Institute: An AI Research Institutes program supported by NSF under Award No. 2019897, and the Institute for Geospatial Understanding through an Integrative Discovery Environment (I-GUIDE) by NSF under Award No. 2118329, US DARPA INCAS Program No. HR0011-21-C0165 and BRIES Program No. HR0011-24-3-0325.



\bibliographystyle{ACM-Reference-Format}
\balance

\bibliography{ref}


\end{document}

%% file: sections/0-abstract.tex
\begin{abstract}
Large Language Models (LLMs) have revolutionized natural language processing with their remarkable capabilities in text generation and reasoning. However, these models face critical challenges when deployed in real-world applications, including hallucination generation, outdated knowledge, and limited domain expertise. Retrieval And Structuring (RAS) Augmented Generation addresses these limitations by integrating dynamic information retrieval with structured knowledge representations.
This survey (1) examines retrieval mechanisms including sparse, dense, and hybrid approaches for accessing external knowledge; (2) explore text structuring techniques such as taxonomy construction, hierarchical classification, and information extraction that transform unstructured text into organized representations; and (3) investigate how these structured representations integrate with LLMs through prompt-based methods, reasoning frameworks, and knowledge embedding techniques. 
It also identifies technical challenges in retrieval efficiency, structure quality, and knowledge integration, while highlighting research opportunities in multimodal retrieval, cross-lingual structures, and interactive systems. This comprehensive overview provides researchers and practitioners with insights into RAS methods, applications, and future directions.
\footnote[0]{* P. Jiang and S. Ouyang contributed equally.}
\end{abstract}

%% file: sections/1-introduction.tex
\section{Introduction}

Large Language Models (LLMs) have revolutionized natural language processing, demonstrating unprecedented capabilities in tasks ranging from text generation to complex reasoning~\cite{brown2020language, achiam2023gpt}. These models, trained on vast amounts of text data, have shown remarkable abilities in understanding context, generating human-like responses, and adapting to various tasks with minimal instruction~\cite{wei2022chain, ouyang2022training}. However, when deployed in real-world applications, LLMs face several critical challenges: they can generate plausible but factually incorrect information (hallucination)~\cite{ji2023survey, huang2025survey}, rely on potentially outdated training data~\cite{lewis2020retrieval, mialon2023augmented}, and show limited expertise in specialized domains~\cite{singhal2023large, pan2024unifying}.

The limitations of LLMs become particularly apparent in knowledge-intensive applications where accuracy and reliability are paramount. While these models excel at pattern recognition and text generation, they struggle with maintaining factual consistency and accessing current information. In specialized fields such as scientific research, healthcare, or technical domains, LLMs often lack the precise and detailed knowledge required for reliable performance~\cite{liu2023evaluating, jiang2025reasoningenhanced, edge2024local}.

To address these challenges, Retrieval-Augmented Generation (RAG) was proposed as a framework that enhances LLM capabilities by retrieving relevant information from external knowledge sources before generating responses~\cite{lewis2020retrieval, gao2023retrieval}. By grounding LLM responses in retrieved documents, RAG reduces hallucinations and enables access to up-to-date information not present in the model's parameters. Despite these advancements, traditional RAG approaches face fundamental limitations: they typically process unstructured text passages without leveraging structured knowledge, contain non-atomic information that can mislead LLMs~\cite{edge2024local, wu2024medical, jiang2025reasoningenhanced}, and struggle with complex queries requiring multi-hop reasoning or domain-specific knowledge organization~\cite{zhong2023comprehensive, zhao2024retrieval}.

Retrieval And Structuring (RAS) has emerged as a more powerful paradigm that addresses these limitations by integrating dynamic information retrieval with structured knowledge representations. While RAG provides the foundation for connecting LLMs with external information, RAS extends this capability by incorporating knowledge structuring techniques that transform unstructured text into organized representations such as taxonomies, hierarchies, and knowledge graphs~\cite{shen2018hiexpan, zhang2024teleclass, ding2024automated}. These structured representations help organize retrieved information, guide the retrieval process, and provide a framework for verifying LLM outputs.

Recent advances in RAS have demonstrated significant improvements in LLM performance through novel retrieval strategies that better align with LLM reasoning patterns~\cite{gao2021complement, wang2022neural} and improved methods for constructing high-quality knowledge representations~\cite{komarlu2023ontotype, jiang2025kg}. The integration of retrieval and structuring mechanisms has opened new possibilities across domains: in scientific research, RAS systems synthesize information while maintaining technical accuracy~\cite{dagdelen2024structured, yasunaga2022deep}; in e-commerce, they generate personalized recommendations~\cite{chen2023knowledge, yu2022folkscope}; and in healthcare, they provide reliable contextualized information~\cite{karim2023large, wen2023mindmap}.

Several challenges remain in fully realizing RAS potential: efficient retrieval at scale~\cite{thakur2021beir, zhu2023large}, maintaining high-quality knowledge representations~\cite{zhong2023comprehensive}, seamlessly integrating structured information with LLM reasoning~\cite{sun2024thinkongraph, luo2023chatrule}, and balancing computational overhead with real-time performance~\cite{li2023can, dong2023self}.

This survey provides a comprehensive examination of RAS, its components, applications, and future directions. We begin with preliminaries on LLMs and retrieval-augmented generation, then examine advanced retrieval mechanisms and text structuring approaches, including taxonomy construction, hierarchical classification, and information extraction. We analyze how these structured representations integrate with LLMs to improve reasoning and adaptation capabilities, examine domain-specific applications, and conclude with technical challenges and research opportunities.

%% file: sections/2-background.tex



\section{Preliminaries and Foundations}

This section provides a concise overview of the preliminaries and foundational aspects of Large Language Models (LLMs) and Retrieval-Augmented Generation (RAG).

\subsection{Large Language Models}

\subsubsection{Architectures}

LLMs are primarily built upon three architectural frameworks:
\textbf{(1) Encoder-only Models}, like BERT~\citep{devlin2019bert} and its derivatives, designed for understanding tasks, process input text to generate embeddings or predict class labels, excelling in applications such as text classification and information extraction.
\textbf{(2) Encoder-Decoder Models}, like BART~\citep{bart} and T5~\citep{raffel2020exploring}, consist of both encoder and decoder components, enabling them to transform input text into output text. This architecture is particularly effective for natural language generation tasks like machine translation and text summarization.
\textbf{(3) Decoder-only Models}, like GPT~\citep{achiam2023gpt} and DeepSeek~\citep{deepseek_v3}, are optimized for generative tasks. By predicting subsequent tokens in a sequence, these models excel in open-ended applications like text completion and dialogue generation, establishing their dominance in the generative AI landscape.

\subsubsection{Training Paradigms}
The development of LLMs involves several key stages:
\textbf{(1) Pre-training.} Initially, models are exposed to extensive text corpora to learn language patterns through tasks like Masked Language Modeling~\citep{devlin2019bert} and Autoregressive Language Modeling~\citep{gpt1}, building a foundational understanding of linguistic structures.
\textbf{(2) Supervised Fine-Tuning (SFT).} After pre-training, LLMs can undergo instruction tuning to better follow human instructions rather than merely predicting the next tokens~\citep{instructgpt}. This process trains models on instructional prompts and responses, enhancing their ability to interpret and execute user requests. Additionally, models can be fine-tuned either from a pre-trained state or an instruction-tuned version to optimize performance for specific tasks.
\textbf{(3) Reinforcement Learning (RL).} Reinforcement learning techniques refine model behavior by aligning outputs with human preferences using methods like Proximal Policy Optimization (PPO)~\citep{ppo} and Direct Preference Optimization (DPO)~\citep{dpo}. Although RL typically follows SFT stage, it can also be applied directly to pre-trained models, as shown in DeepSeek-R1-Zero~\citep{deepseek_r1}.

\subsubsection{Inference Strategies}

Modern LLMs leverage advanced inference techniques to enhance performance and scalability: \textbf{(1) In-Context Learning (ICL).} ICL~\citep{gpt3} enables models to adapt to new tasks by incorporating examples directly into the input prompt, facilitating task-specific responses without additional training.
\textbf{(2) Chain-of-Thought (CoT).} CoT prompting~\citep{wei2022chain} guides models to generate intermediate reasoning steps, boosting performance on complex tasks by making the problem-solving process explicit.
\textbf{(3) Test-Time Scaling.} Recent work, such as OpenAI's o1~\citep{openai_o1}, has explored strategies to improve reasoning capabilities by allocating more computational resources during inference. This approach enables models to perform more extensive deliberation, like longer CoT reasoning, before generating responses. By scaling computation effort at test time, models can achieve substantial improvements on complex tasks, particularly in science and mathematics.

\subsection{Retrieval-Augmented Generation}
Retrieval-Augmented Generation (RAG) enhances LLMs by retrieving relevant information from external sources before generating responses \cite{lewis2020retrieval}. RAG addresses hallucination, outdated knowledge, and domain expertise limitations by grounding responses in retrieved documents \cite{ji2023survey, huang2025survey}. The RAG pipeline includes indexing (document processing/vectorization), retrieval (identifying relevant information), and generation (with retrieved context) \cite{borgeaud2022improving, gao2023retrieval}.

\subsubsection{RAG Paradigms}

\smallskip \noindent\textbf{(1) Naive RAG.} This initial approach employs a straightforward "retrieve-read" framework \cite{ma2023query}, where documents are retrieved based on the original query and directly fed into the model. While effective, this approach struggles with retrieval precision and may introduce irrelevant information that can mislead LLMs \cite{chen2024benchmarking, wu2024medical, jiang2025reasoningenhanced}. \textbf{(2) Advanced RAG.} This paradigm introduces pre-retrieval and post-retrieval optimizations. Pre-retrieval techniques focus on improving indexing structures and query formulation through query rewriting \cite{ma2023query}, query expansion \cite{peng2024large}, and hybrid retrieval approaches \cite{gao2021complement}. Post-retrieval processes enhance integration through reranking and context compression techniques \cite{blagojevi2023enhancing}, significantly improving response relevance.
\textbf{(3) Modular RAG.} The most recent paradigm represents a flexible architecture incorporating specialized components for specific tasks \cite{wang2023knowledgpt}. This approach introduces functional modules like search interfaces, memory systems, and task adapters, enabling sophisticated retrieval strategies \cite{kim2023tree} and patterns like rewrite-retrieve-read \cite{ma2023query}, generate-read \cite{yu2022generate}, and demonstrate-search-predict \cite{khattab2022demonstrate}.

\subsubsection{Key Components and Techniques}

\smallskip \noindent\textbf{(1) Retrieval Sources.} While unstructured text data remains predominant \cite{wang2022training}, research explores semi-structured data \cite{li2023table}, structured knowledge graphs \cite{wang2023knowledgpt, kang2023knowledge}, and even LLM-generated content \cite{cheng2023lift, sun2022recitation}.
\textbf{(2) Retrieval Granularity and Embeddings.} Granularity ranges from fine-grained (tokens, sentences) to coarse-grained (chunks, documents) units \cite{chen2024dense}, each balancing information completeness against retrieval precision. Embedding models have seen substantial improvements through specialized architectures \cite{li2023angle, nussbaum2024nomic} and alignment fine-tuning \cite{zhang2023retrieve, dai2022promptagator}.
\textbf{(3) Context Processing.} Reranking methods \cite{zhuang2023open} prioritize the most relevant documents while context compression techniques \cite{xu2023recomp} reduce redundancy while preserving essential content, addressing the "lost in the middle" problem \cite{liu2024lost}.

\subsubsection{Advanced Workflows and Integration}
 RAG has evolved beyond static pipelines to include adaptive retrieval \cite{jiang2023active, asai2023self}, where models autonomously determine when retrieval is necessary; iterative retrieval \cite{shao2023enhancing}, which employs recursive cycles to refine responses; and recursive retrieval \cite{kim2023tree}, which hierarchically decomposes complex queries.
Despite increasing context windows in modern LLMs, RAG maintains importance through efficiency, transparency, and domain adaptability \cite{packer2023memgpt, xiao2023efficient}. The approach continues to evolve through integration with fine-tuning \cite{lin2023ra, luo2023augmented}, creating hybrid systems. Recent evaluations reveal that while fine-tuning shows some improvement, RAG consistently outperforms it for both existing and new knowledge \cite{ovadia2023fine}. Research has also extended RAG to incorporate other modalities like images \cite{yasunaga2022retrieval} and code \cite{li2023structure}.



%% file: sections/3-retrieval.tex
\section{Information Retrieval}
Information Retrieval (IR)~\citep{hambarde2023information, faloutsos1995survey, mitra2000information} aims to identify and retrieve relevant documents from a corpus based on queries. It plays a crucial role in various tasks including internet browsing~\citep{kobayashi2000information, pokorny2004web, lewandowski2005web}, question-answering systems~\citep{karpukhin2020dense, thakur2021beir, abbasiantaeb2021text}, and LLM applications~\citep{gao2023retrieval, zhao2024retrieval, fan2024survey}. In the RAS paradigm, IR serves as a critical component that provides external, up-to-date, yet unstructured knowledge. The quality of the IR process directly impacts the overall effectiveness of the whole framework. \label{sec:retrieval}

\subsection{Classical Retrieval Methods}

\subsubsection{Sparse Retrieval}
Sparse retrieval has long been the de facto standard for information retrieval~\citep{hambarde2023information, guo2022semantic, luan2021sparse}. At the core, it represents text using normalized term frequency and inverse document frequency. 

\smallskip
\noindent\textbf{Fundamental Approaches.} 
A fundamental approach in this category is TF-IDF~\citep{sparck1972statistical}, which scores terms based on their frequency in a document while penalizing common words that appear across many documents. BM25~\citep{robertson1995okapi, robertson2009probabilistic}, an improvement over TF-IDF, refines this approach by introducing a saturation function for term frequency and document length normalization.

\smallskip
\noindent\textbf{Statistical Modifications.} 
Several statistical variations of BM25 have been introduced~\citep{trotman2014improvements}. BM25F~\citep{robertson2009probabilistic, perez2010using, shi2014empirical} extends BM25 for fielded search by considering term weights across multiple document fields. BM25+~\citep{lv2011lower} introduces different frequency saturation mechanisms. Other variants include BM25-adpt~\citep{Lv2011Adaptive}, LM-DS~\citep{Zhai2001Smoothing}, and LM-PYP~\citep{Momtazi2010HPYP}, each optimizing the BM25 formula for different retrieval scenarios.

\smallskip
\noindent\textbf{Neural Modifications.} 
Learned sparse retrieval incorporates neural networks to generate sparse, term-based representations~\citep{dai2019context, lin2021few, formal2021splade, mallia2022faster}. DeepCT~\citep{dai2019context} and uniCOIL~\citep{lin2021few} leverage transformers to re-weight terms, assigning weights based on semantic importance. SPLADE~\citep{formal2021splade} utilizes masked language modeling to predict additional relevant terms, blending semantic strengths of deep learning with efficiency of inverted indexes.

\smallskip
\noindent\textbf{Impact.} 
Even in the deep learning era, sparse retrieval remains a strong baseline due to its effectiveness, efficiency and interpretability. It continues to be competitive with more complex neural models, particularly with limited training data or domain-specific constraints~\citep{thakur2021beir}. However, term-based scoring cannot retrieve relevant content without lexical overlap and can miss broader context or meaning~\citep{luan2021sparse, arabzadeh2021predicting}.

\subsubsection{Dense Retrieval}
In contrast to sparse lexical methods, dense retrieval~\citep{zhao2024dense, gao2021condenser, zhu2023large} uses continuous vector embeddings to represent queries and documents. With pre-trained language models~\citep{devlin2019bert, raffel2020exploring, ni2021sentence}, relevant query–document pairs have high cosine similarity, enabling semantic matching without lexical overlap.

\smallskip
\noindent\textbf{Dual Encoder and Early Dense Retrieval.} 
Early dense retrieval emerged as a response to the limitations of sparse retrieval. \citet{karpukhin2020dense} designed Dense Passage Retrieval (DPR), which showed that a BERT~\citep{devlin2019bert}-based dual encoder trained on question–answer pairs~\citep{kwiatkowski-etal-2019-natural, joshi2017triviaqa, berant2013semantic, baudivs2015modeling, rajpurkar2016squad} can surpass BM25~\citep{robertson1995okapi, robertson2009probabilistic} in open-domain question answering. Early models struggled with generalization, requiring large-scale labeled datasets~\citep{nguyen2016ms} and appropriate negative sampling strategies~\citep{xiong2020approximate} in contrastive learning~\citep{faghri2017vse++, oord2018representation, he2020momentum, chen2020simple}. 

\smallskip
\noindent\textbf{Effectively-trained Dense Retrieval.} 
Recognizing limitations of static negative sampling, Asynchronous Negative Contrastive Estimation (ANCE)~\citep{xiong2020approximate} addressed this by refreshing hard negatives during training. RocketQA~\citep{qu2020rocketqa} improved contrastive training with cross-batch memory for more diverse negative samples. Generalizable T5 Retriever (GTR)~\citep{ni2021large} framed retrieval as a sequence-to-sequence task rather than solely relying on contrastive learning.

\smallskip
\noindent\textbf{Efficient and Scalable Dense Retrieval.} 
Another research line focused on scalability and efficiency. ColBERT~\citep{khattab2020colbert} introduced late interaction mechanisms, preserving token-level representations to reduce memory overhead. CoCondenser~\citep{gao2021unsupervised} optimized intermediate representation learning for more compact document embeddings. E5~\citep{wang2022text} explored improved contrastive pretraining techniques for adaptability to various tasks.

\smallskip
\noindent\textbf{Impact.} 
Dense Retrieval models~\citep{karpukhin2020dense, xiong2020approximate, ni2021large, khattab2020colbert, wang2022text} offer significant advantages over Sparse Retrieval by enabling semantic matching when exact lexical overlap is absent. However, Dense Retrieval is limited by dependency on large-scale labeled training data~\citep{fang2024scaling, zhao2024dense, li2024domain} and lack of interpretability~\citep{cheng2024interpreting, llordes2023explain, kang2024interpret}.

\subsubsection{Others}
In addition to sparse and dense retrieval, alternative methods exist for retrieving relevant text.

\smallskip
\noindent\textbf{Retrieval with Edit Distance.} 
Edit distance-based Retrieval matches queries and documents by comparing edit distances between words and phrases like Levenshtein Distance~\citep{yujian2007normalized, heeringa2004measuring, haldar2011levenshtein}. These methods are especially popular in code retrieval~\citep{hayati2018retrieval}, often integrated with abstract syntax trees of code snippets~\citep{zhang2020retrieval, poesia2022synchromesh}.

\smallskip
\noindent\textbf{Retrieval with Named Entity Recognition.} 
Named Entity Recognition (NER)~\citep{nadeau2007survey, yadav2019survey, li2020survey}-based retrieval uses entities in queries and documents for matching, allowing text to be understood at a concept level~\citep{xu2016fofe, lin2020bridging}. Entity-based retrieval enforces type matching, especially useful for specific scenarios like legal domain~\citep{dozier2010named, leitner2019fine}.

\smallskip
\noindent\textbf{Retrieval with Large Language Models.} 
Recently, Generative Retrieval (GR)~\citep{tay2022transformer, wang2022neural, sun2023learning, de2020autoregressive, bevilacqua2022autoregressive} has emerged as a new paradigm leveraging generative model~\citep{raffel2020exploring, touvron2023llama, achiam2023gpt} parameters for memorizing documents, enabling retrieval by directly generating relevant document identifiers. Given the same parameter size and training data, GR can achieve better performance~\citep{tay2022transformer, wang2022neural} and interpretability~\citep{tay2022transformer, zhang2024generative}, becoming a new research focus~\citep{li2024matching}.

\subsection{Recent Retrieval Strategies}
While building on major retrieval frameworks discussed above, diverse strategies have been recently introduced at various stages of the retrieval pipeline.

\subsubsection{Hybrid Retrieval}
Hybrid retrieval~\citep{gao2021complement, luan2021sparse, mandikal2024sparse} combines different retrieval models to leverage complementary strengths. \citet{gao2021complement} proposed CLEAR (Complementing Lexical Retrieval with Semantic Residual), learning a dense "residual" vector to complement BM25. \citet{luan2021sparse} explicitly combined sparse and dense representations, achieving better ranking accuracy than strong single-method baselines. This fused retrieval fashion enhances effectiveness and robustness~\citep{tu2022mia, zhang2024efficient}.

\subsubsection{Data Augmentation for Retrieval}
To enlarge training scale, data augmentation~\citep{nogueira2019document, nogueira2019doc2query, bonifacio2022inpars, chaudhary2023s} leverages additional data beyond original labeled queries and documents. Doc2query~\citep{nogueira2019document} trained a transformer~\citep{vaswani2017attention} to predict likely queries for each document. DocT5query~\citep{nogueira2019doc2query} used T5~\citep{raffel2020exploring} to generate multiple queries per document. Inpars~\citep{bonifacio2022inpars} used GPT-3~\citep{brown2020language} to generate queries for documents in target domains. These techniques inject additional knowledge into retrieval systems, becoming key drivers of performance improvement.

\subsubsection{Query Expansion and Rewriting}
Query expansion~\citep{roy2016using, naseri2021ceqe, 10.1145/2071389.2071390, AZAD20191698} and rewriting~\citep{ma2023query, liu2024query, ye2023enhancing} aim to improve retrieval by reformulating the input query. \citet{roy2016using} expanded queries with semantically related terms using word2vec~\citep{mikolov2013efficient}; \citet{naseri2021ceqe} applied BERT to contextualize expansion based on retrieved documents. \citet{ma2023query} trained LMs to rewrite queries using initial retrieval outputs. \citet{jiang2025deepretrieval} introduced \textit{DeepRetrieval}, an RL-based approach that optimizes retrieval metrics (e.g., Recall@K, NDCG) directly without supervision, and achieves SOTA performance across literature, evidence-seeking, and SQL retrieval tasks with only 3B parameters.
Experimental results show that effective query rewriting/expansion significantly improves retrieval quality. 
Building on this line, \textit{s3}~\citep{jiang2025s3} further demonstrates that training a search-only agent with reinforcement learning—using a generation-aware reward rather than search-only metrics—can yield strong downstream improvements while requiring orders of magnitude less training data.

\subsubsection{Multi-staged Retrieval and Reranking}
For efficiency and effectiveness, reranking processes~\citep{nogueira2019passage, gao2024llm, kudo2024document, chuang2023expand, chen2020simple} adopt two-stage retrieval strategies, where documents from a fast method are re-scored by a more powerful but slower ranking model. \citet{nogueira2019passage} demonstrated BERT-based rerankers dramatically improve ranking quality. As NLP advances, LLMs~\citep{raffel2020exploring, touvron2023llama, brown2020language} have emerged as state-of-the-art rerankers, boosting retrieval performance by large margins on both general~\citep{gao2024llm, kudo2024document, chuang2023expand, chen2020simple} and specific domains~\citep{ajith2024litsearch}.

%% file: sections/4-structuring.tex
\section{Text Structuring}
Text structuring is essential for converting unstructured text into organized, interpretable knowledge for analysis and decision-making. This section explores three key text mining methodologies: taxonomy construction and enrichment, text classification, and information extraction. These approaches effectively organize textual information, enabling downstream tasks like knowledge graph construction, database population, and tabular relational learning.

\subsection{Text Mining}

\subsubsection{Taxonomy}
Taxonomy is a tree-like structure that organizes concepts into a hierarchical framework, where parent nodes represent broader categories, and child nodes capture finer-grained subcategories. Taxonomy is often used as the label space structuring method for text-mining tasks.

\smallskip
\noindent\textbf{Taxonomy Construction.}
Taxonomy construction typically begins with a seed structure. HiExpan~\cite{shen2018hiexpan} expands entity taxonomies through width and depth expansions, followed by global structure adjustment using word analogy~\cite{10.1145/3178876.3186024}. CoRel~\cite{Huang2020CoRel} leverages pre-trained language models (PLMs) to train a relation-transferring module that generalizes seed parent-child relationships. TaxoCom~\cite{Lee2022TaxoComTT} completes partial topical taxonomies by learning local discriminative word embeddings and applying novelty-adaptive clustering.

A related task is set expansion, which extends a seed set with semantically similar entities. SetExpan~\cite{shen2017setexpan} uses skip-grams and word embeddings to measure similarity between candidates and the seed set. Set-CoExpan~\cite{huang2020setcoepan} introduces auxiliary sets through embedding learning. CGExpan~\cite{zhang2020empower} exploits PLMs to generate knowledge-probing queries, while ProbExpan~\cite{li2022probexpan} refines entity representations using contrastive learning. FGExpan~\cite{Xiao2023FGExpan} focuses on fine-grained set expansion, targeting specific common types.

\smallskip
\noindent\textbf{Taxonomy Enrichment.}
Taxonomy enrichment enhances each taxonomy node with discriminative texts, such as keywords. It can be classified into flat and hierarchical enrichment. Flat enrichment methods often use PLM embeddings for clustering~\cite{zhang2022neural, meng2022topic} or discriminative topic modeling. CatE~\cite{Meng2020DiscriminativeTM} creates a joint word embedding space, while KeyETM~\cite{Harandizadeh2022KeywordAE} extends embedded topic modeling~\cite{Dieng2020TopicMI} by integrating topical keywords. GTM~\cite{churchill2022guided} focuses on guided topic modeling for short text. SeeTopic~\cite{zhang2022seed} addresses unseen topic names by leveraging PLMs' general knowledge. SeedTopicMine~\cite{Zhang2023SeedTopicMine} explores diverse context information for topic modeling.
Hierarchical enrichment requires modeling additional structure information. TaxoGen~\cite{zhang2018taxogen} recursively clusters word embeddings and refines them using local corpora. NetTaxo~\cite{shang2020nettaxo} extends this by incorporating network structure information. JoSH~\cite{Meng2020HierarchicalTM} leverages a taxonomy to train a joint embedding space that preserves relative tree distances between nodes.

\subsubsection{Text Classification}

Text classification categorizes text into different categories. According to how the label space is organized, it can be divided into flat classification (independent categories) and hierarchical classification (tree-structured labels).

\smallskip
\noindent\textbf{Flat Classification.}
Flat classification has been extensively studied. Early works~\cite{yang-etal-2016-hierarchical, zhang2015character} focused on designing task-specific objectives for different architectures. Later, external supervision sources such as knowledge bases~\cite{Song2014OnDH}, human-curated rules~\cite{ratner2016, badene-etal-2019-data, chatterjee2019data}, and keywords~\cite{meng2018weakly, mekala-shang-2020-contextualized, ren2020denoisingMS, onan2016ensemble} were utilized. Keywords inspired self-training methods like WeSTClass~\cite{meng2018weakly}, LOTClass~\cite{meng-etal-2020-text}, and X-Class~\cite{wang-etal-2021-x}. ClassKG~\cite{zhang-etal-2021-weakly} combined keywords and knowledge bases by constructing a keyword graph using GNNs.

With the rise of LLMs, prompting-based methods~\cite{zhu2023prompt, wen2023augmenting, han2022ptr, wang2021transprompt} achieved notable results by leveraging LLMs' internal knowledge. \citet{min2021noisy} proposed channel models for few-shot learning, estimating conditional probabilities. NPPrompt~\cite{zhao-etal-2023-pre} introduced a zero-shot approach that builds class verbalizers using PLM embeddings. PIEClass~\cite{zhang-etal-2023-pieclass} explored discriminative PLM prompting for pseudo-labeling. CARP~\cite{sun-etal-2023-text} leveraged LLMs’ reasoning abilities for zero and few-shot classification, using CoT prompting to identify indicative clues before labeling.

\smallskip
\noindent\textbf{Hierarchical Classification.} 
Hierarchical text classification assigns input text to one or more nodes within a taxonomy-structured label space, presenting a greater challenge than flat classification due to its extensive structure. Most studies adopt either fully-supervised or semi-supervised approaches. \citet{banerjee-etal-2019-hierarchical} and \citet{wehrmann2018hierarchical} train multiple classifiers for each node or level of the taxonomy. Others~\cite{peng2018deepgraphcnn, chen-etal-2021-hierarchy, wang-etal-2022-incorporating, jiang-etal-2022-exploiting, wang-etal-2023-towards-better} train a unified classifier for the entire taxonomy. However, these methods heavily rely on manual annotation, making weakly-supervised approaches increasingly popular. WeSHClass~\cite{meng2019weakly} extends WeSTClass~\cite{meng2018weakly} by modeling label hierarchies as mixtures of distributions and using self-training. HiMeCat~\cite{zhang2021hierarchical} integrates metadata by learning joint representations for labels, metadata, and text using a hierarchical generative model, which supports weak supervision through augmented documents. TaxoClass~\cite{shen-etal-2021-taxoclass} leverages a pre-trained entailment model and employs taxonomy-guided search for pseudo-training samples. TELEClass~\cite{zhang2024teleclass} enhances taxonomies with class-indicative features, adapting LLMs for hierarchical classification by selecting pseudo labels from retrieved candidates and generating pseudo documents for long-tail and fine-grained classes.

\subsubsection{Information Extraction}\label{section413}
Information extraction is another important task for text mining, focusing on entity-level information as building blocks for knowledge structuring.

\smallskip
\noindent\textbf{Entity Mining.}
Entity mining includes named entity recognition (NER) and fine-grained entity typing (FET). NER involves extracting entities from text before recognition. Early approaches addressed uncertainty in distantly supervised labels: AutoNER's "Tie or Break" strategy~\cite{shang2018learning}, PaTTER's dictionary expansion~\cite{9378052}, and ETAL's confident entity identification~\cite{chaudhary-etal-2019-little}. RoSTER~\cite{meng-etal-2021-distantly} addressed incomplete labels using noise-robust learning. SDNET~\cite{chen-etal-2022-shot} pre-trained on Wikipedia entities. SEE-Few~\cite{yang-etal-2022-see} utilized an entailment framework for learning from limited examples. Recent LLM approaches include GPT-NER~\cite{wang2023gpt} with self-verification, \citet{shang2025local}'s document graphs, and X-NER~\cite{peng-etal-2023-less} for extremely weakly supervised settings.
\
FET deals with a more detailed label space organized into an ontology. ZOE~\cite{zhou2019zero} utilized a taxonomy based on Freebase types. OntoType~\cite{komarlu2023ontotype} aligned fine-grained types with an ontology using weak supervision. OnEFET~\cite{ouyang2023ontology} enriched ontology structures with instances. ALIGNIE~\cite{10.1145/3534678.3539443} employed prompt-based approaches for interpreting entity types. SEType~\cite{zhang2024seed} incorporated additional entities from unlabeled corpora.
\
Ultra-fine entity typing (UFET)~\cite{choi-etal-2018-ultra} expanded the label space to thousands of types. BERT-MLMET\cite{dai2021ultra} fine-tuned BERT with supervision from headwords and hypernyms. LITE~\cite{li2022ultra} leveraged natural language inference for entity typing. DenoiseFET~\cite{li2023ultra} clustered the extensive label space into centroids for better generalization.

\smallskip
\noindent\textbf{Relation Extraction.}
Building on mined entities, relation extraction (RE) identifies relationships between entity pairs, playing a crucial role in knowledge graph construction (Sect. \ref{sec:knowledge_structuring}) and event induction~\cite{kargupta2024unsupervised, DBLP:conf/emnlp/ShenZJ021, DBLP:conf/www/JiaoZSZZ023}. Leveraging LLMs, \citet{wadhwa-etal-2023-revisiting} demonstrate that few-shot prompting in GPT-3 can rival state-of-the-art fully supervised RE models and explore LLM-based data augmentation with CoT explanations. \citet{wan-etal-2023-gpt} enhance in-context learning by utilizing task-aware sentence-level and entity-level representations for kNN retrieval of examples.
RelationPrompt~\cite{DBLP:conf/acl/ChiaBPS22} prompts LLMs to synthesize relation examples for generating structured texts. To minimize hallucinations and enhance factual accuracy, STAR~\cite{DBLP:journals/corr/abs-2305-15090} and DocGNRE~\cite{DBLP:conf/emnlp/LiJZ23} adopt self-refinement and self-validation strategies. REPaL~\cite{zhou2024grasping} suggests that relation descriptions offer broader semantic coverage than mere relation names. However, LLMs show limitations in relational reasoning, possibly due to the scarcity of RE tasks in instruction-tuning datasets~\cite{wang-etal-2022-super, zhang-etal-2023-aligning}. Addressing this, \citet{ma2023large} propose the LLM-SLM cooperation framework, a \textit{filter-then-rerank} approach, showcasing LLMs' effectiveness in information extraction. QA4RE~\cite{zhang-etal-2023-aligning} reframes RE as a question-answering task to align with instruction-tuning, while SumAsk~\cite{li2023revisiting} combines summarization and QA. GenRES~\cite{jiang-etal-2024-genres} criticizes traditional evaluation metrics like precision and recall for generative RE and introduces a comprehensive multi-dimensional evaluation framework.
Additionally, \citet{obamuyide-vlachos-2018-zero} and \citet{sainz-etal-2021-label} reformulate RE as a natural language inference (NLI) task. REBEL~\cite{huguet-cabot-navigli-2021-rebel-relation} redefines relation triplet extraction as a Seq2seq task using decoder-based LMs, while SURE~\cite{lu2022summarization} adopts a summarization approach by verbalizing relation names. Prompt-tuning models~\cite{han2022ptr, 10.1145/3485447.3511998} also improve relational reasoning by integrating diverse prompts.
Another research direction focuses on aggregating relation-indicative evidence. REPEL~\cite{10.1145/3178876.3186024} co-trains a pattern-based module with a distributional module for RE, while RClus~\cite{10.1007/978-3-031-43421-1_2} leverages patterns of relation-specific entity types and indicative words in dependency paths. Eider~\cite{xie-etal-2022-eider} adopts a document-level approach to extract sentence-level evidence, and SAIS~\cite{xiao-etal-2022-sais} supervises intermediate steps like coreference resolution and entity typing to retrieve fine-grained, relation-specific evidence.

\subsection{Knowledge Structuring}\label{sec:knowledge_structuring}
Structured knowledge, such as knowledge graphs, databases, and tabular data, has long been studied in knowledge structuring~\cite{zhong2023comprehensive}.

\smallskip
\noindent\textbf{Knowledge Graph Construction.} 
Knowledge graphs (KGs) represent real-world entities as nodes and their relationships as edges.
\textit{In the general domain}, open information extraction (OpenIE) tools extract triples from raw documents, which are then filtered, linked, and merged to form KGs. However, pipelined methods often suffer from error propagation, leading to the development of end-to-end approaches. For instance, \citet{zeng-etal-2018-extracting} propose a sequence-to-sequence model that jointly extracts entities and relations, addressing the overlap problem through a copy mechanism. Similarly, frameworks like REBEL~\cite{huguet-cabot-navigli-2021-rebel-relation} and ABSA~\cite{yan-etal-2021-unified} leverage pre-trained language models (PLMs) to translate raw text into structured knowledge. TagReal~\cite{jiang-etal-2023-text} uses pattern mining to generate high-quality prompts for KG completion.
With the rise of large language models (LLMs), new methods for KG construction have emerged. KnowledgeGraph GPT~\cite{kggpt} directly prompts GPT-4 to convert plain text into KGs. AutoKG~\cite{DBLP:journals/www/ZhuWCQOYDCZ24} employs multiple LLM agents to enhance KG construction and reasoning. KG-FIT~\cite{jiang2025kg} refines KG embeddings by constructing coherent entity hierarchies and fine-tuning them using both hierarchical and textual information.
\textit{Domain-specific KG construction} requires substantial human input for annotations and ontology development. For example, \citet{rotmensch2017learning} integrates annotated medical records with existing KGs to build a health-related KG. PAIR~\cite{gan2023making} uses LLMs as relation filters to streamline relation discovery in online marketing. In the biomedical field, \citet{karim2023large} build ontologies to validate gene-disease relationships and use BioBERT to generate RDF triples, later refined by LLMs. MedKG~\cite{jiangmedkg} emphasizes human-in-the-loop interaction by allowing users to modify nodes and edges. In e-commerce, FolkScope~\cite{yu-etal-2023-folkscope} adopts a semi-automated approach using domain-specific prompts to guide LLMs, while \citet{chen2023knowledge} investigates prompting strategies for effective KG construction in few-shot settings.
\textit{Theme-specific KG construction} is more challenging for its focus on specialized or fast-evolving domains. TKGCon~\cite{DBLP:journals/corr/abs-2404-19146} addresses this by starting with a theme-specific entity ontology from Wikipedia and using LLMs to generate relation ontologies. Entities and relations are mapped to the corpus to form the final KG.

\smallskip
\noindent\textbf{Database Population.} 
Databases are structured systems with executable operations to manage data~\cite{DBLP:journals/corr/abs-2402-01763}. AVATAR~\cite{jayram2006avatar} established probabilistic database techniques for IE. \citet{li2023can} examines LLMs as database interfaces. DB-GPT~\cite{zhou2024db} proposes an LLM-based framework for database-specific tasks. D-Bot~\cite{DBLP:journals/pvldb/ZhouLSLCWLFZ24} automatically diagnoses databases. ChatDB~\cite{DBLP:journals/corr/abs-2306-03901} augments LLMs with SQL databases for complex reasoning. Text2DB~\cite{jiao2024text2db} designs an agent system to automatically populate databases from text.

\smallskip
\noindent\textbf{Tabular Data Organization.}
Tables are a prevalent structured format for various applications. RolePred \cite{jiao2022open} utilizes information redundancy in multiple documents to extract structured tables including the attribute names and the corresponding values of an event type. ODIE \cite{jiao2023instruct} adopts instruction tuning for large language models to extract a single table from the texts following on-demand user instructions. In particular, the table structures can be defined by users or automatically inferred by the model. TableLLAMA \cite{zhang2023tablellama} and Table-GPT \cite{li2023table} also fine-tune large language models with table-related data to improve their ability to process tables, like row population, column addition, and cell filling. StructGPT~\cite{DBLP:conf/emnlp/JiangZDYZW23} further design specialized interfaces to interact with tables for reasoning tasks. EVAPORATE~\cite{arora2023language} transforms semi-structured documents into tabular views. RelBench~\cite{fey2023relational} reformulates relational databases as heterogeneous graphs for representation learning with GNNs.

%% file: sections/5-ras-app.tex





\section{Retrieval-And-Structuring}
The Retrieval-And-Structuring (RAS) paradigm unifies three complementary approaches to knowledge processing: information retrieval, structured knowledge representation through Knowledge Graphs (KGs), and Large Language Models (LLMs). This integrated framework enhances LLMs' ability to perform knowledge-intensive tasks, enables more effective knowledge discovery within structured knowledge, and creates synergistic benefits across all three components. In this section, we examine the foundational principles of this paradigm, analyze recent methodological advances, and explore promising directions for future research.

\begin{figure}[t]
    \begin{center}
    \includegraphics[width=0.8\columnwidth]{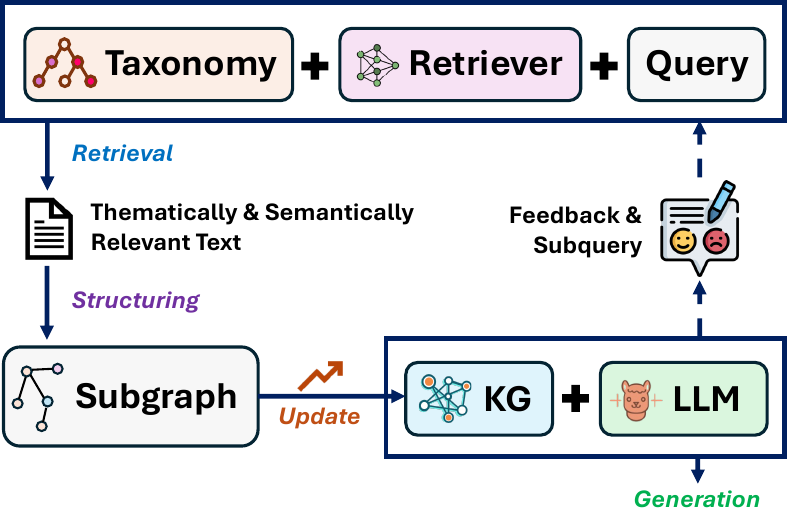}
    \end{center}
    \vspace{-3mm}
    \caption{An abstractive example of RAS paradigm. }
    \label{fig:ras_abstract}
    \vspace{-1em}
\end{figure}

\subsection{Structure-Enhanced Retrieval}
\noindent\textbf{Taxonomy-based approaches}. Classic retrieval methods \cite{salton1988term, robertson2009probabilistic, formal2021splade, karpukhin2020dense, khattab2020colbert, izacard2021unsupervised}, while effective for general search, are not optimized for theme-specific searches with explicit thematic focus.
In contrast, taxonomy-guided retrieval utilizes the thematic structures captured by the domain-specific topic taxonomy structures, which does not only align closely with the topics and subtopics in a taxonomy but also bypasses the requirement for large-scale labeled/annotated data. 
Along this track, ToTER~\cite{kang2024improving} augments dense retrieval by leveraging corpus taxonomies through three strategies: search space filtering via topic overlap, relevance matching through topic distributions, and query enrichment using taxonomy-guided phrases, which allows improved domain-specific search without requiring labeled data.
TaxoIndex~\cite{kang-etal-2024-taxonomy} extracts key concepts from papers and organizes them as a semantic index guided by academic taxonomies at both topic and phrase levels, then trains an add-on module to identify these concepts in queries and documents to improve matching of underlying academic concepts in dense retrievers, achieving strong performance even with limited training data. In a similar vein, Lima et al.~\cite{teixeira-de-lima-etal-2025-know} demonstrates that retrieval performance varies significantly across different query taxonomy classes, reinforcing the importance of structured categorization for optimizing retrieval.

\noindent\textbf{KG-based approaches}. Knowledge graph, as another structured data, also demonstrated its effectiveness to improve the retrieval quality. For instance, KG-RAG~\cite{sanmartin2024kg} reduces LLM hallucinations~\cite{ji2023survey} by constructing KGs from text and retrieving information through a Chain of Explorations algorithm, which sequentially traverses the graph using LLM reasoning to find relevant answers with higher precision than vector-based retrieval. Xu et al.~\cite{xu2024retrieval} leverage KGs for retrieval by parsing issue tickets into tree structures with interconnected relationships, then using entity matching and graph traversal to locate relevant information instead of relying on text-chunk similarity search. HippoRAG~\cite{gutierrez2024hipporag} constructs a KG from passages using LLM-based triple extraction, then leverages personalized PageRank~\cite{10.1145/511446.511513} to explore graph neighborhoods around query entities to identify relevant passages with implicit connections.
Similarly, GFM-RAG~\cite{luo2025gfm} follows this KG construction approach but innovates in the retrieval mechanism by employing a pre-trained query-dependent graph foundation model, enabling more sophisticated multi-hop relationship reasoning in a single step.

Inspired by the recent advancements in retrieval strategies introduced in Section \ref{sec:retrieval}, we depict a promising RAS method in Figure~\ref{fig:ras_abstract}.
The method combines taxonomy and KG to enable precise knowledge retrieval. The process follows an iterative cycle where a taxonomy-enhanced retriever first identifies both thematically and semantically relevant documents from the corpus. These documents are structured into a subgraph, which updates a query-specific evolving KG. The KG augments the LLM's capabilities by providing essential, structured information for answering the query. If additional information is needed, the LLM generates a focused subquery conditioned on the KG, initiating another retrieval cycle until sufficient knowledge is gathered. This feedback loop ensures comprehensive and accurate responses by dynamically building and refining a knowledge context tailored to the query.

\subsection{Structure-Enhanced LLM Generation.}
Once knowledge is structured into a comprehensive KG, the challenge shifts to leveraging this representation to enhance LLM outputs. Structure-enhanced LLM generation grounds model responses in explicit knowledge structures, reducing hallucinations and improving factual consistency.

\subsubsection{Early Approaches (PLM + KG)}
Early integration approaches broadly fall into neural and rule-based categories:
\noindent\textbf{(1) Neural Integration.} Initial neural methods fused KGs with pre-trained language models through various architectural innovations. KG-BART \cite{liu2021kg} augmented the BART architecture with graph attention networks to inject knowledge for commonsense reasoning. JointLK~\cite{sun2021jointlk} proposed a joint reasoning framework connecting language models and KGs through a shared semantic space. Wang et al.~\cite{wang2023unifying} unified structure reasoning with pre-training by creating cross-modal representations. GreasLM~\cite{zhang2022greaselm} enhanced KG integration by introducing graph-aware semantic aggregation that improved question answering over heterogeneous knowledge sources. Dragon~\cite{yasunaga2022deep} took a different approach with deep bidirectional language-knowledge graph pre-training that created entity-aware representations.
\noindent\textbf{(2) Rule-based Integration.} Complementing neural approaches, rule-based methods used explicit reasoning paths over KGs. MHGRN~\cite{feng2020scalable} implemented multi-hop graph relation networks that performed explicit message passing between entities relevant to the query. QA-GNN~\cite{yasunaga2021qa} addressed this by jointly reasoning over language and KG representations, formulating paths through a graph neural network architecture. UniKGQA~\cite{jiang2023unikgqa} unified retrieval and reasoning through transformers operating directly on knowledge subgraphs, demonstrating better generalization to novel domains and question types.
These early approaches laid the foundation for more sophisticated integration of structured knowledge with language models, establishing key mechanisms for knowledge grounding and explicit reasoning paths.

\subsubsection{Modern Approaches (LLM + KG)}
Recent advancements showcase stronger integration strategies between LLMs and KGs, moving beyond early fusion methods to develop more sophisticated reasoning frameworks. These approaches enable LLMs to leverage the structured representations in KGs more effectively, improving reasoning capabilities while maintaining interpretability.

\noindent\textbf{LLM reasoning over KG structure.} ToG~\cite{sun2024thinkongraph} enables LLMs to reason directly with KGs through guided exploration. Starting from key entities in the query, the model navigates the KG step-by-step using prompt-driven reasoning to build explicit reasoning chains. ToG-2~\cite{ma2025thinkongraph} tightly couples KG navigation with document contexts through bidirectional iterative ``knowledge-guided context retrieval'' and ``context-enhanced graph retrieval'', enabling more precise reasoning. Besta et al.~\cite{Besta_Blach_Kubicek_Gerstenberger_Podstawski_Gianinazzi_Gajda_Lehmann_Niewiadomski_Nyczyk_Hoefler_2024} introduce Graph-of-Thought (GoT), which structures LLM reasoning as a graph rather than a linear chain. GoT enables multi-path exploration, backtracking, and cyclical thinking by representing reasoning steps as nodes and their relationships as edges. Graph CoT~\cite{jin2024graph} augments LLMs with graphs through an iterative framework of reasoning, interaction, and execution. The model identifies information needs, generates graph function calls, and processes retrieved data sequentially, enabling step-by-step traversal of complex graphs. RoG~\cite{luo2024reasoning} introduces a planning-retrieval-reasoning framework where LLMs first generate relation paths grounded in KGs as plans, then retrieve valid reasoning paths for structured reasoning. ChatRule~\cite{luo2023chatrule} utilizes LLMs to mine logical rules for KG reasoning, combining semantic understanding with KG structure. It samples paths from KGs to prompt LLMs for rule generation, ranks rules using confidence metrics, and applies them for reasoning tasks without extensive rule enumeration, preserving interpretability. MindMap~\cite{wen2023mindmap} enables LLMs to construct mind maps from KG inputs, combining path-based and neighbor-based evidence subgraphs into reasoning graphs. The approach helps LLMs perform synergistic inference by merging retrieved KG facts with implicit knowledge, reducing hallucinations while maintaining transparent reasoning pathways, particularly effective in medical domains. ORT~\cite{liu2025ontology} introduces ontology-guided reverse thinking for KGQA, constructing reasoning paths from query targets back to conditions using KG ontology. The approach extracts both condition and target entities, builds backward reasoning trees, and prunes irrelevant paths to guide precise knowledge retrieval. 

\noindent\textbf{KG-Embedded LLM} is another track of research that aims to directly incorporate KG information into the LLM's architecture or training process, rather than using KGs as external resources during inference time. This approach allows for more seamless integration of structured knowledge within the generation process.
GraphToken \cite{fatemi2024talk} introduces a parameter-efficient method to encode structured graph data for LLMs. Unlike text-based approaches, GraphToken learns an encoding function that generates soft-token prompts to extend textual input with explicit structural information. The system employs a graph neural network encoder to capture graph structure while a projection layer aligns these representations with the LLM's embedding space. By freezing the LLM parameters and only training the graph encoder, GraphToken provides a computationally efficient way to enhance LLMs' graph reasoning capabilities.
Building upon GraphToken, G-Retriever \cite{he2025g} extends this approach by incorporating retrieval-augmented generation for improved textual graph understanding. While maintaining the graph encoder architecture to align representations with the LLM's embedding space, G-Retriever addresses scaling challenges by formulating subgraph retrieval as a Prize-Collecting Steiner Tree optimization problem. This allows the system to select only the most relevant portions of large graphs, enabling natural language interaction with graphs through a conversational interface while reducing hallucination in the generation.

\noindent\textbf{Structure Summarization} is another way to leverage knowledge structures for LLM enhancement by condensing complex structured knowledge into text formats.
GraphRAG~\cite{edge2024local} organizes knowledge through hierarchical community detection, creating multi-level summaries of graph communities. Its map-reduce approach generates partial answers from relevant community summaries before aggregating them into comprehensive responses, effectively handling queries that require global corpus understanding.
KARE~\cite{jiang2025reasoningenhanced} employs selective community retrieval using a multi-faceted relevance scoring system that considers node overlap, coherence, recency, and thematic relevance. Its dynamic retrieval algorithm prioritizes diverse, valuable communities to augment contexts, enabling LLMs to generate more accurate predictions and interpretable reasoning.

%% file: sections/6-future.tex
\section{Future Directions}

\subsection{Technical Challenges}

\subsubsection{Retrieval Efficiency} 
Ensuring fast and accurate retrieval of relevant information remains challenging as LLMs and data repositories grow. Future work requires developing scalable indexing methods for high-throughput requests while maintaining low latency \cite{lewis2021paq,izacard2023atlas}. Adaptive retrieval strategies—tailoring search depth based on query complexity or user context—can enhance performance and user satisfaction \cite{achiam2023gpt}. Addressing these needs holistically is essential for real-world applications demanding both speed and accuracy.

\subsubsection{Knowledge Quality} 
The quality of structured representations is critical, yet automation of taxonomy construction and information organization can introduce noise or inconsistencies. Future research should focus on robust validation techniques and iterative refinement incorporating domain expertise \cite{PaLM2Blog2023}. Leveraging contextual signals—discourse patterns, domain ontologies, user feedback—can guide more fine-grained structuring, ensuring extracted knowledge remains coherent across domains.

\subsubsection{Integration Challenges} 
Integrating disparate knowledge sources poses a key challenge for effective RAS systems. This process involves reconciling conflicts, minimizing redundancy, and updating representations as new information emerges. Incremental learning and conflict-resolution strategies are crucial for maintaining consistency as knowledge evolves \cite{pan2024unifying}. Building adaptive frameworks that can ingest and reconcile varied data sources in real time will better equip LLMs to provide accurate, context-rich answers.

\subsection{Research Opportunities}

\subsubsection{Multi-modal Knowledge Integration} 
As data becomes increasingly diverse, future RAS solutions must handle images, video, and audio alongside text. Unified multi-modal indexing strategies capturing cross-modal semantic relationships can significantly enhance information discovery \cite{liang2024survey}. Neural architectures combining vision-language models with text encoders can facilitate cross-modal reasoning, allowing LLMs to reference non-textual cues in responses.

\subsubsection{Cross-lingual Systems} 
Techniques leveraging multilingual embeddings and parallel corpora can build language-agnostic taxonomies enabling global knowledge sharing. Transfer learning methods reusing structured representations from high-resource languages can accelerate development in low-resource languages \cite{achiam2023gpt}. Standardizing cross-lingual benchmarks with robust domain adaptation will drive accuracy across multilingual contexts\cite{chirkova2024retrieval}.

\subsubsection{Interactive and Self-Refining Systems} 
Conversational interfaces allow users to refine queries naturally, leading to more precise retrieval. Self-refinement mechanisms using reinforcement learning or meta-learning enable systems to identify and correct errors autonomously, mitigating hallucinations and outdated knowledge \cite{dong2023self}. This combined approach improves both accuracy and explainability as models document their corrective steps while adapting to user feedback\cite{asai2023self, shao2023enhancing}.

\subsubsection{Human-AI Collaborative Frameworks} 
While automation is central to RAS, human expertise remains essential for complex tasks. Expert reviewers can guide taxonomy refinement, while crowd-sourcing methods help maintain data relevance over time \cite{mosqueira2023human}. These collaborative frameworks ensure models adhere to privacy regulations and fairness principles, fostering trustworthy solutions that leverage both machine scalability and human judgment\cite{ouyang2022training}.

\subsubsection{Personalized Knowledge Delivery} 
Tailoring retrieval and structuring processes to individual users significantly enhances relevance. Personalized approaches incorporate user profiles, historical behavior, or explicit feedback to shape retrieval strategies \cite{liu2004personalized}, while privacy-preserving techniques safeguard personal data. By layering contextual cues on core architectures, these systems can deliver insights that evolve with user needs, providing more engaging interactions \cite{geng2022improving, jiang2023graphcare}.

%% file: sections/7-conclusion.tex
\section{Conclusions}
This survey presented Retrieval-And-Structuring (RAS) as a paradigm that enhances LLMs by integrating dynamic retrieval with structured knowledge to mitigate hallucinations, outdated information, and limited domain expertise. We reviewed key techniques in retrieval, text structuring, and their integration with LLMs for knowledge-grounded generation. Despite progress, challenges remain in retrieval efficiency, knowledge quality, and integration. Promising directions include multimodal and cross-lingual systems, self-refining frameworks, human-AI collaboration, and personalized knowledge delivery—advancing LLM reasoning with factual and current knowledge across domains.